\pdfoutput=1
\documentclass[11pt]{article}

\usepackage[final]{acl}
\usepackage{times}
\usepackage{latexsym}
\usepackage{ragged2e}

\usepackage[T1]{fontenc}
\usepackage[utf8]{inputenc}

\usepackage{microtype}

\usepackage{inconsolata}

\usepackage{graphicx}

\usepackage[T1]{fontenc}
\usepackage[utf8]{inputenc}
\usepackage{amsmath}
\usepackage{amssymb}
\usepackage{microtype}
\usepackage{booktabs}
\usepackage{multirow}
\usepackage{subcaption} % For subfigure support
\usepackage{float} % For [H]
\usepackage{enumitem}
\usepackage{latexsym}
\usepackage{inconsolata} % For monospace font
\usepackage{setspace}
\usepackage{titlesec} % For section spacing adjustments
\usepackage{color}
\usepackage{colortbl}
\usepackage{cite}
\usepackage{fontawesome5}
\usepackage{xcolor}
\usepackage{pgfplotstable}
\usepackage{soul}
\usepackage{tcolorbox}
\usepackage{longtable}

\usepackage{xcolor}

\author{
{\bf Hadi Mohammadi}\textsuperscript{1}\thanks{These authors contributed equally to this work.}\thanks{Corresponding author: \texttt{h.mohammadi@uu.nl}}\quad
{\bf Tina Shahedi}\textsuperscript{1}\footnotemark[1]\quad
{\bf Pablo Mosteiro}\textsuperscript{1}\\
{\bf Massimo Poesio}\textsuperscript{2,3}%
\quad
{\bf Ayoub Bagheri}\textsuperscript{1}%
\quad
{\bf Anastasia Giachanou}\textsuperscript{1}\\[6pt]
\textsuperscript{1}\small{Department of Methodology and Statistics, 
Utrecht University, The Netherlands} \\
\textsuperscript{2}\small{Department of Information and Computing Sciences, 
Utrecht University, The Netherlands} \\
\textsuperscript{3}\small{Queen Mary University of London, London, United Kingdom
}}

\pgfplotsset{compat=1.18}
\usepackage{amssymb}  % For \checkmark and \xmark 

\tcbset{
    warningbox/.style={
        colback=orange!10!white, colframe=orange!80!black,
        fonttitle=\bfseries, coltitle=orange!50!black,
        enhanced, sharp corners, boxrule=0.5mm, width=\textwidth,
        before=\par\smallskip, after=\par\smallskip, left=5pt, right=5pt, top=5pt, bottom=5pt
    }
}

\title{Assessing the Reliability of LLMs Annotations\\
in the Context of Demographic Bias and Model Explanation}
\begin{document}
\maketitle
%Warning text with proper width control
\begin{minipage}{\textwidth}
\scriptsize
{\raggedright
\color{orange} \faExclamationTriangle \, The paper contains examples which are offensive in nature.\par}
\end{minipage}

\begin{abstract}

Understanding the sources of variability in annotations is crucial for developing fair NLP systems, especially for tasks like sexism detection where demographic bias is a concern. This study investigates the extent to which annotator demographic features influence labeling decisions compared to text content. Using a Generalized Linear Mixed Model, we quantify this influence, finding that while statistically present, demographic factors account for a minor fraction (~8\%) of the observed variance, with tweet content being the dominant factor. We then assess the reliability of Generative AI (GenAI) models as annotators, specifically evaluating if guiding them with demographic personas improves alignment with human judgments. Our results indicate that simplistic persona prompting often fails to enhance, and sometimes degrades, performance compared to baseline models. Furthermore, explainable AI (XAI) techniques reveal that model predictions rely heavily on content-specific tokens related to sexism, rather than correlates of demographic characteristics. We argue that focusing on content-driven explanations and robust annotation protocols offers a more reliable path towards fairness than potentially persona simulation. 

\end{abstract}

%Key words
%Sexism Detection, Demographic Bias, LLM Annotations, Persona Prompting, Explainable AI

\section{Introduction}

\begin{justify}
Reliable annotations are foundational to machine learning in NLP, guiding models toward accurate predictions. According to~\citet{uma2020case}, annotation involves humans labeling or transforming data inputs into "gold data", which guides machine learning practitioners in building their models. For instance, to create a gold dataset for a model that corrects grammatical errors, annotators might be asked to identify mistakes in a range of sample sentences. However, creating high-quality annotations
\end{justify}

\begin{figure}[H]
\centering
\includegraphics[width=0.4\textwidth]{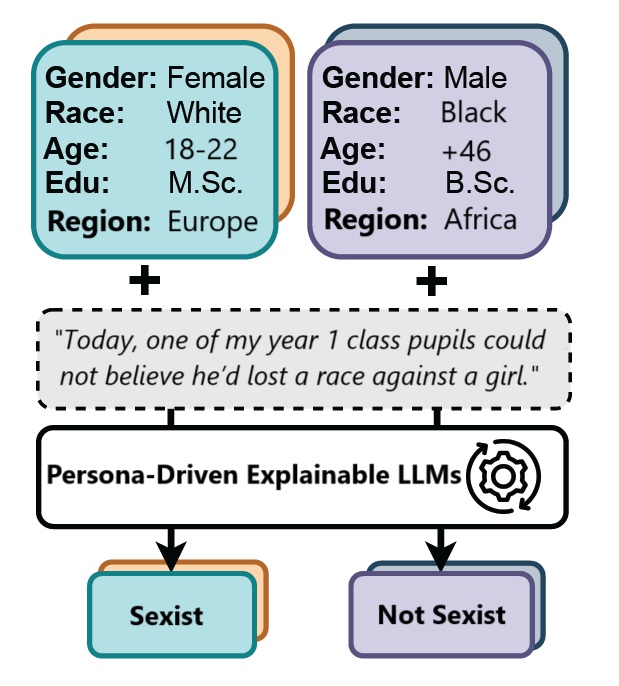}
\caption{\small We instruct LLMs to replicate human annotations for subjective NLP tasks from different perspectives using persona prompting and XAI techniques. Our results show that simulated personas alone may not sufficiently capture human subjectivity. XAI analysis confirms that tweet content plays a more significant role in model decisions.}
\label{fig:Persona-Driven Explainable LLMs}
\end{figure}
\vspace{-12pt}

\noindent is not a straightforward task since it requires thoughtful consideration of the criteria that make annotations effective, consistent, and unbiased. 

This raises the following question, what defines a robust annotation process? When it comes to evaluating annotation quality, several studies highlight Inter-Annotator Agreement (IAA), as defined by \citet{krippendorff2011computing}, as a standard metric for labeled datasets~\citep{pei2023annotator, plank2014learning}. However, achieving high IAA is often challenging, particularly for subjective language tasks that rely on human judgments. For tasks like sexism detection, where subjectivity is inherent, addressing annotator agreement challenges is essential, as disagreements can significantly influence the performance of NLP models trained on this data. In some cases, disagreement often arises from ambiguous sentences or vague label definitions, which can make it difficult for annotators to reach an agreement~\citep{russell2008labelme, artstein2008inter}. Traditionally, aggregating judgments from multiple annotators to create a single "ground truth" for each data instance is widely used to address the inherent ambiguity and subjectivity in language interpretation. This approach is similar to initial methods for handling annotator disagreement, which focuses on estimating a "true" label. However,~\citet{pavlick2019inherent} shows that even when annotators are provided with additional context, there is not always a single correct answer, and disagreements still persist. 

Recent studies indicate a significant shift in how annotator disagreements are handled, particularly in subjective tasks involving human judgments~\citep{pavlick2019inherent, basile2021we, uma2021learning, plank2022problem}. Current research primarily focuses on developing models that can learn from these disagreements. While NLP researchers aim for consistency among annotators, some level of disagreement is both inherent and unavoidable in human annotation processes~\citep{leonardelli2021agreeing}. As~\citet{bless2014social} showed, annotators’ demographic factors, personal perspectives, and differing value systems can lead to discrepancies in annotations. 

Building on this foundation, researchers have systematically analyzed how the characteristics of annotators and the way tasks are framed can skew evaluation outcomes. For instance,~\citet{hoskinghuman} demonstrate that annotator assertiveness and the linguistic complexity of model outputs significantly bias judgments of factuality and consistency in crowdsourced error annotations. Their study finds that responses that sound more confident are judged as more accurate, even if they contain the same number of errors. Similarly,~\citet{kirk2024prism} reveal that factors such as cultural background, age, gender, and personal values lead to substantial variation in how responses are rated for helpfulness, creativity, and alignment with individual beliefs. These findings underscore the challenge of distinguishing true model performance from annotator-induced biases and motivate the need for more scalable and consistent annotation methods under controlled conditions.

This sparked researchers to explore the potential of GenAI models as substitutes for human annotators. Several studies have shown that large language models (LLMs), when provided with demographic information, can imitate specific annotator groups by tailoring their outputs to reflect attributes such as gender, race, age, or education ~\citep{beck2024sensitivity, schafer2024demographics}. However, LLMs often align more closely with certain demographics (e.g., younger, White, male) unless explicitly directed otherwise ~\citep{schafer2024demographics}. To the best of our knowledge, the incorporation of XAI techniques to guide these models is still rare ~\citep{ralevski2024using, he2024annollm, ArgumentativeLLMs}. This creates critical gaps in evaluating how demographic biases impact annotation reliability and whether GenAI models, with XAI guidance, can effectively substitute human annotators, especially in subjective tasks such as sexism detection. For instance,~\citet{mohammadi2024transparent} presents an explainability-enhanced sexism detection pipeline that bridges model predictions with token-level explanations, illustrating efforts to improve transparency in sexism detection.

In this study, we use data from the EXIST 2024 challenge~\citep{plaza2024exist}, a shared task on sexism detection in social networks, a highly subjective task.\footnote{\scriptsize\url{https://nlp.uned.es/exist2024/}} Our primary goal is to assess annotation reliability and examine how demographic biases influence annotator decisions. Using a Generalized Linear Mixed Model (GLMM), we analyze both fixed and random effects, revealing that demographic variables account for nearly 8\% of the variance in labeling behavior, suggesting the presence of demographic biases in human judgments.We also evaluate LLM performance by simulating annotation and classification under various prompting scenarios, model configurations, and temperature settings. Our methodology compares state-of-the-art models across open-source frameworks and proprietary APIs, exploring how prompt modifications affect outcomes. To improve explainability, we employ SHAP values to reveal the influence of specific tokens on predictions across demographic groups. By integrating SHAP analyses into persona prompting, we examine how demographic attributes shape predictions. Results show that combining SHAP with persona prompting enhances both interpretability and reliability of LLM-generated annotations.

\paragraph{Bias Statement}
This paper examines how demographic factors, such as gender, ethnicity, education, and region, may influence both human and LLM annotations in detecting sexist content on social media. We focus on potential representational harms, wherein certain demographic groups’ viewpoints or sensitivities to biased language might be underrepresented or misjudged. By highlighting differences in labeling behaviors across diverse annotator backgrounds, we aim to reduce the risk that an NLP system trained on these annotations will inadvertently perpetuate stereotypes or unfairly discount certain cultural experiences. We take the normative stance that all groups deserve unbiased and respectful treatment in both data collection and model predictions. Our ultimate goal is to ensure that technology, especially in sensitive tasks like sexism detection, does not exacerbate inequalities or reinforce harmful narratives.

\section{Related Work}
\label{Related_Work}

Recent studies have explored how annotators' personal backgrounds, experiences, and identities influence labeling outcomes, particularly in subjective tasks \citep{pei2023annotator}. However, findings in this area are mixed. Some studies report significant correlations between demographic features and annotation results \citep{excell2021towards}, while others observe minimal statistically significant differences, especially regarding gender \citep{biester2022analyzing}. These conflicting results highlight the complexity of the relationship between annotator characteristics and labeling decisions. Contrasts are particularly evident in tasks such as identifying sexist content, offensive language, and political ideologies, where an individual's personal experiences and group affiliations can significantly influence their perception and categorization of content \citep{kamruzzaman2024woman}. The diversity of findings underscores the need for ongoing research to better understand the intricate interplay between annotator attributes and labeling outcomes. This understanding is crucial for developing more robust and inclusive NLP models that can effectively incorporate diverse perspectives in the annotation process. While some studies attempt to enhance data quality by analyzing disagreements among annotators, systematic investigations into how annotators’ demographic biases affect annotation results remain limited \citep{gupta2023sociodemographic}. 

\subsection{Persona Prompting for LLMs Annotations} 
One promising approach to NLP annotation tasks involves using GenAI Models, such as GPT-4, which have been explored for automating annotation tasks due to their advanced language understanding capabilities~\citep{manikandan2023language}. Furthermore, LLMs have shown potential in simulating diverse human perspectives by integrating demographic features into prompts \citep{hu2024quantifying}. This technique, known as "persona prompting", has been effectively utilized to model human behavior and facilitate role-playing scenarios \citep{beck2024sensitivity}. For instance \citet{hu2024quantifying} examined how demographic, social, and behavioral persona variables influence LLM predictions and highlighted the importance of considering personal attributes in subjective NLP tasks. The success of LLMs in this domain has sparked discussions about their potential to replace human subjects in research contexts, particularly in annotation tasks \citep{dillion2023can, grossmann2023ai}. 

However, this raises concerns about identity misrepresentation and the flattening of group nuances \citep{wang2024large}. Moreover, persona prompting is not without its challenges. LLMs may carry inherent biases from their training data, potentially affecting annotation quality \citep{bender2021dangers, pavlovic2024effectiveness}. Recent studies highlight these limitations, noting that LLMs often replicate societal biases or fail to adequately capture the nuances of minority perspectives \citep{hu2025generative, pavlovic2024effectiveness}. These issues emphasize the need for nuanced techniques to evaluate and mitigate the extent to which LLMs can accurately simulate human-like predictions. 

\subsection{LLMs Annotations' Interpretability} 
XAI techniques, particularly SHAP (SHapley Additive exPlanations), have become powerful tools for improving model interpretability by attributing importance to input features~\citep{zhao2024explainability}. In NLP, SHAP effectively identifies influential tokens driving classification decisions and uncovers potential model biases~\citep{ribeiro2016should}. Recent advances have expanded XAI's role in tasks such as sentiment classification, bias detection, toxic language identification, and inference~\citep{he2024annollm}. \citet{he2024annollm} introduced a two-step framework using GPT-3.5, where the model first generates explanations and then annotates data through prompting. This approach has achieved performance comparable to or exceeding human annotators in tasks like Question Answering (QA) and Word-in-Context (WiC), demonstrating the potential of LLMs for annotation. Similarly, \citet{ralevski2024using} applied GPT-3.5 and GPT-4 for annotating housing instability using chain-of-thought prompting. While LLMs are not yet suitable for full automation due to challenges such as bias, they show strong potential for computer-assisted annotation, reducing the time and cost of manual efforts.

\section{Experimental Setup}
% \subsection{Generalized Linear Mixed Model}
\label{sec:Experimental_Setup}

% \paragraph{Dataset}
\vspace{-3pt}
\subsection{Dataset}
\label{sec:EXIST_2024_data}

We used data from the EXIST 2024 challenge~\citep{plaza2024exist}, which comprises datasets sourced from Twitter (now X). The labeled dataset contains tweets in both English and Spanish, with the training set comprising 6920 tweets in both languages (3260 in English, and 3660 in Spanish). For simplicity, we focus exclusively on Task 1 which involves binary classification of tweets to determine whether they express content related to sexism. Each tweet in the dataset was annotated by six individuals, who also provided demographic information across five categories: gender, age, ethnicity, education, and country. Specifically, gender was recorded as male or female; age was grouped into three categories (18–22, 23–45, and 46+); ethnicity included Asian, Black, White, Latino, Middle Eastern, Multiracial, and Other; education levels ranged from less than high school to doctorate; and annotators came from 45 countries. To simplify the analysis, these countries were categorized into five regions: Europe, America, Africa, Asia, and the Middle East. This grouping reduced the total number of unique demographic combinations from 266 to 117. We then eliminated combinations with rare representations, which we explain in detail in the next section.

% \paragraph{Generalized Linear Mixed Model}
\subsection{Generalized Linear Mixed Model}
\label{subsection: Generalized Linear Mixed Model}
We ran a GLMM to examine how annotators’ demographic features affect labeling decisions. The model accounts for clustering of labels within tweets by incorporating random effects, ensuring that demographic influences are estimated independently of tweet-specific characteristics and individual differences. In our dataset, tweets and annotators serve as grouping variables, forming a crossed random effects structure: each tweet is labeled by multiple annotators, and each annotator labels multiple tweets. Also, tweets are hierarchically nested within languages. To account for both crossed and nested random effects, the following mixed-effects logistic regression model is specified.\footnote{\scriptsize{\textbf{In R notation, } $\texttt{label}_{ij} \sim \texttt{Annotators' demographic factors} + (1 \mid \texttt{lang/id\_EXIST}) + (1 \mid \texttt{annotator\_id})$}}

{\footnotesize
\vspace{-3pt}
\[
E(\text{label}_{ij} \mid \mathbf{b}) = \text{logit}^{-1}(\mathbf{X}_{ij}\boldsymbol{\beta} + \mathbf{Z}_{ij}\mathbf{b})
\]
}
\vspace{-15pt}

In the model, \( \text{label}_{ij} \) is the binary response variable indicating whether the label for the \( i \)-th tweet by the \( j \)-th annotator is \texttt{YES} or \texttt{NO}. The design matrix \( \mathbf{X}_{ij} \) includes fixed effects for annotator demographic features, with \( \boldsymbol{\beta} \) representing their corresponding coefficients. Random effects are modeled as \( \mathbf{Z}_{ij}\mathbf{b} \), capturing variation among tweets nested within languages and annotators. The random effects vector \( \mathbf{b} \) follows a multivariate normal distribution \( \sim N(0, \mathbf{G}) \). A logistic inverse link function, \( \text{logit}^{-1}(\cdot) \), is used to model the binary outcome. This model evaluates demographic biases while accounting for tweet-level variability and annotator differences. Following prior studies \citep{pei2023annotator}, we excluded rare demographic features (i.e., representing less than 2\% of annotators), such as the “Middle Eastern” ethnicity with only three annotators. Consequently, 69 out of 725 annotators were removed. We also excluded unique demographic combinations represented by only one annotator unless present in both languages. This resulted in 56 unique demographic combinations, detailed in Appendix \ref{appendix:demographiccombinations}, Table \ref{tab:unique_combinations}.To address demographic and label-class imbalances, we assigned weights to each observation based on the inverse frequency of its demographic attributes and label class. The raw weight (\( W_{\text{raw}} \)) for each observation was calculated as:

{\footnotesize
\vspace{3pt}
$ W_{\text{raw}} = \prod_{\text{features}} \frac{1}{f_{\text{group}}} \times \frac{1}{f_{\text{label}}} $
\vspace{3pt}
}

Here, \( f_{\text{group}} \) denotes the relative frequency of a demographic category, and \( f_{\text{label}} \) the label class frequency. This approach, commonly used in survey weighting to address sample imbalances \citep{groves2009survey}. For computational stability, raw weights were normalized to [0, 1] using \( W_{\text{norm}} = \frac{W_{\text{raw}}}{\max(W_{\text{raw}})} \) and then scaled for use in the mixed-effects model. As shown in Appendix~\ref{appendix:demographic_combination}, Figure \ref{fig:weight_contribution}, the top ten demographic combinations with the highest weight contributions are identified across both \texttt{YES} and \texttt{NO} labels. For instance, female annotators aged 23--45, identifying as Black, holding a bachelor’s degree, and residing in Africa, provide the most balanced weighted input.

% \paragraph{BERT Model and SHAP Values} 
\subsection{BERT Model and SHAP Values}
\label{subsection: BERT Model and SHAP Values}
To classify texts as sexist or non-sexist, we use the Bidirectional Encoder Representations from Transformers (BERT) multilingual model. BERT captures word context by considering both left and right surroundings in a sentence~\citep{devlin2018bert}. The multilingual version is particularly suited to our dataset, which contains texts in two languages. During training, we fine-tune the BERT multilingual model using standard procedures. We use the Adam optimizer with a learning rate of \(3 \times 10^{-5}\) and a batch size of 128. The maximum sequence length is set to 512 tokens to handle longer texts. Binary cross-entropy is used as the loss function for this binary classification task. The model is trained for up to 10 epochs, with early stopping based on validation loss to prevent overfitting and ensure good generalization~\citep{brownlee2018gentle}. To incorporate explainability into our methodology, we use SHAP values, following the approach by~\citep{mohammadi2024transparent}. SHAP values quantify each token's contribution to the model’s prediction, highlighting the most influential parts of the text. The SHAP value for each token \( t \), \( S_t \), is computed by measuring the change in the model’s output when the token is included versus omitted across all possible subsets of input tokens. The SHAP value \( S_t \) for token \( t \) is computed as:

\begin{scriptsize}

% \label{eq:shap}
\vspace{3pt}
$ S_t = \sum_{T' \subseteq T \setminus \{t\}} \frac{|T'|! \left( |T| - |T'| - 1 \right)!}{|T|!} \left[ f(T' \cup \{t\}) - f(T') \right]$
\end{scriptsize}
\vspace{3pt}

Where \( T \) is the set of all tokens in the input text, \( T' \) is a subset of \( T \) excluding token \( t \), and \( f(\cdot) \) represents the model's prediction function. To find the most influential tokens, we calculate the SHAP importance \( \text{SI}_t \) for each token \( t \) by averaging the absolute SHAP values across all instances \( N_t \) where the token appears, considering only the cases where the model's prediction matches the true label:

{\footnotesize
% \label{eq:average_shap}
\vspace{3pt}
$\text{SI}_t = \frac{1}{N_t} \sum_{i=1}^{N_t} \left| S_t(i) \right| \cdot \mathbb{I}\left( y_i = \hat{y}_i \right) $
}
\vspace{3pt}

Here, \( S_t(i) \) is the SHAP value of token \( t \) in instance \( i \), \( y_i \) is the true label, \( \hat{y}_i \) is the predicted label, and \( \mathbb{I}(\cdot) \) is the indicator function. After that, we normalize the SHAP importance scores to compute the importance ratio for each token: 
{\footnotesize
$\text{IR}_t = \frac{\text{SI}_t}{\sum_{k \in T} \text{SI}_k}.$
}
Tokens are ranked by importance ratios, and cumulative importance is calculated as 
{\footnotesize
$\text{CI}_k = \sum_{i=1}^{k} \text{IR}_i$
}
to select the most influential tokens such that \( \text{CI}_k \leq T_c \). We set the threshold \( T_c = 0.95 \) to retain tokens contributing to 95\% of the total importance. These top tokens are identified per class and incorporated into the GenAI prompts by bolding them, guiding the generative model to focus on critical parts of the text. Integrating SHAP enhances classifier transparency, revealing key factors driving decisions. Crucially, analyzing high-importance tokens helps determine whether the model relies on meaningful indicators of sexism or spurious correlations. These tokens are then used in GenXAI and GenPXAI scenarios, which will be described in more detail in the section~\ref{subsec:GenAI Scenarios}, to guide LLMs, allowing us to assess whether highlighting content-relevant features improves annotation reliability.

\vspace{-3.9pt}

\subsection{Large Language Models}
\label{subsec: Large Language Models}
We experiment with a range of LLMs, including local open-source models and cloud-based proprietary APIs, including OpenAI-based models (GPT-4o and GPT-4o mini)\footnote{\scriptsize\url{https://openai.com/index/hello-gpt-4o/} \& \scriptsize\url{https://openai.com/index/gpt-4o-mini-advancing-cost-efficient-intelligence/}} and LLaMA-based models (LLaMA 3.2 3B and LLaMA 3.3 70B).\footnote{\scriptsize\url{https://huggingface.co/meta-llama/Llama-3.2-3B} \& \scriptsize\url{https://huggingface.co/meta-llama/Llama-3.3-70B-Instruct}} We evaluate how factors such as model size and architecture (e.g., OpenAI vs LLaMA variants) influence sexism detection across different scenarios and prompts.

\vspace{-4.5pt}
\subsection{GenAI Scenarios}
\label{subsec:GenAI Scenarios}
We evaluate four main scenarios, each designed to probe the effect of additional context or highlighting on the model's responses. These scenarios, inspired by a previous study~\citep{kamruzzaman2024woman}. The first scenario, \textbf{GenAI}, involves using a generative model to classify texts without additional context or guidance. The prompt instructs the model to make a straightforward classification based solely on the input text.  In the second scenario, \textbf{Persona-Driven GenAI (GenP)}, a persona is added to the prompt to assess its impact on model performance. The persona provides specific demographic characteristics the model should adopt, aiming to influence its perspective and potentially enhance sensitivity to sexist content.
The third scenario, \textbf{Explainable GenAI (GenXAI)}, incorporates SHAP values to highlight influential parts of the text using bold formatting, guiding the model’s attention to key sections. This tests whether focusing on important tokens improves classification accuracy.
In the fourth scenario, \textbf{Persona-Driven Explainable GenAI (GenPXAI)}, we combine persona guidance with SHAP-based highlighting to assess the joint effect of perspective adoption and attention emphasis on model performance.

\vspace{-3pt}
% \paragraph{Defining the Personas}
\subsection{Personas}
\label{subsec: Personas}

Personas are constructed based on demographic attributes such as gender, age, ethnicity, education, and region, corresponding to the 56 unique demographic combinations listed in Appendix~\ref{appendix:demographiccombinations}, Table~\ref{tab:unique_combinations}.  They shape the perspective from which the model evaluates text, potentially reflecting cultural, linguistic nuance or demographic sensitivities.

\begin{itemize}
\footnotesize
    \item \textbf{English:} "\small \texttt{You are a \{gender\} individual, aged \{age\}, who identifies as \{ethnicity\}, has a \{study\_level\}, and currently resides in \{region\}. You have the cultural and personal background of someone with these demographics.}"
\vspace{-5pt}
    \item \textbf{Spanish:} "\small \texttt{Eres una persona \{gender\}, de \{age\} años, que se identifica como \{ethnicity\}, posee un nivel de estudios \{study\_level\}, y actualmente reside en \{region\}. Tienes el trasfondo cultural y personal de alguien con estas características demográficas.}"
\end{itemize}

\vspace{-12pt}
\subsection{Important Tokens}
\label{subsec: Important Tokens}
For scenarios involving GenXAI and GenPXAI, we rely on previously computed important tokens from SHAP values. We highlight the top tokens by wrapping them in bold formatting (\textbf{**token**}) to draw the model's attention. This approach aims to help the model focus on terms that are most indicative of sexism.

\vspace{-3pt}

\subsection{Majority Voting}
\label{subsec: Majority Voting}

Majority voting is used to assign hard labels, while probabilities are used for soft labels. This provides a robust benchmark for evaluating automated methods. To simulate multiple annotators, the model generates six responses per text under each scenario and temperature setting. These six outputs represent "virtual annotators," and majority voting is applied to produce a single prediction per text. This simulates inter-annotator variability and offers a more robust estimate of the model's stance, similar to human annotation aggregation.

\section{Results and Discussion}
\label{sec:Results_Discussion}

\paragraph{Do demographic biases mainly drive labeling differences, or does tweet content play a larger role?} To investigate this question, we first fit a flat logistic regression model with annotator demographic features as fixed effects. This provides a baseline assessment of demographic influence without accounting for tweet-specific or annotator-level variability. We then extend the analysis using a mixed-effects logistic regression model, incorporating crossed random intercepts for annotators and nested random effects for tweets within languages. This approach captures both annotator variability and tweet-specific differences while retaining demographic features as fixed effects. 

Our findings show that incorporating tweet-level and annotator-level variability in the mixed-effects model substantially improves performance over the flat model. The mixed model achieves higher accuracy (73.73\% vs 48.76\%) and F1 score (75.77\% vs 45.09\%), along with better fit indicated by lower AIC and BIC values and a higher AUC. A kappa value of (47.06\%) and an intraclass correlation coefficient (ICC) of 92.3\% highlight the importance of accounting for tweet-specific differences, which the flat model ignores. Notably, the random effect for tweets shows high variance (33.72), indicating that tweet content is the main source of labeling variability. The annotator random effect (5.54) also contributes meaningfully, while the language effect (0.30) has minimal influence. These findings confirm the mixed-effects model as a more accurate and nuanced approach for understanding the labeling process.

\begin{table}[H]
\centering
\scriptsize
\renewcommand{\arraystretch}{0.9}
\setlength{\tabcolsep}{3pt}
\caption{\small Comparison of Flat Model and Mixed-Effects Model Coefficients. Significant codes: ‘***’very strong(\( p < 0.001 \)), ‘**’strong(\( 0.001 \leq p < 0.01 \)), ‘*’moderate(\( 0.01 \leq p < 0.05 \)), ‘\textbf{.}’weak(\( 0.05 \leq p < 0.1 \)), ‘-’very weak(\( 0.1 \leq p < 1 \)).}

%, ‘$\times$’Not significant(\( p \geq 1 \))
\label{tab:Comparison_Flat_Mixed}
\begin{tabular}{p{1.5cm}rrrr}
\toprule
\textbf{Variable} & \textbf{Coef\_Flat} & \textbf{P\_Flat > |z|} & \textbf{Coef\_Mixed} & \textbf{P\_Mixed > |z|} \\
\midrule
(Intercept)$^1$                  &  0.274  & *** & -0.328  & -     \\
Female                      &  \cellcolor[HTML]{e3ffd3} 0.020  & *** &  0.055  & -     \\
23-45                     &  \cellcolor[HTML]{b0ff84} 0.206  & *** &  0.027  & -     \\
46+                       & \cellcolor[HTML]{fcd2d2} -0.089  & *** &  0.111  & -     \\
Black                      & \cellcolor[HTML]{b0ff84} 0.214  & *** & \cellcolor[HTML]{b0ff84} 1.704  & \textbf{.}   \\
Latino                     & \cellcolor[HTML]{fbbaba} -0.237  & *** & \cellcolor[HTML]{f88a8a} -0.770  & *   \\
High school                & \cellcolor[HTML]{fbbaba}-0.255  & *** & \cellcolor[HTML]{faa2a2} -0.465  & *   \\
Master                     & \cellcolor[HTML]{faa2a2}-0.506  & *** &  0.048  & -     \\
Africa                     & \cellcolor[HTML]{f88a8a} -0.732  & *** & \cellcolor[HTML]{f77272} -2.865  & **  \\
America                    & \cellcolor[HTML]{c9ffab} 0.178  &  *** &  0.370  & -     \\
\bottomrule
\end{tabular}

\begin{flushleft}
\scriptsize $^1$ The reference group is male annotators aged 18–22 from Europe who hold a bachelor's degree and identify as white. 

\end{flushleft}
\end{table}
\vspace{-7pt}

Table~\ref{tab:Comparison_Flat_Mixed} compares the coefficients of the flat logistic regression and mixed-effects models for each demographic feature. The flat model assumes independence among observations, ignoring the dataset’s hierarchical structure. As a result, it attributes all variability to fixed effects and residual error, potentially leading to biased coefficient estimates. For example, the flat model suggests females are slightly more likely to label \texttt{YES} than males, but it fails to account for content-specific variability, leading to a misleading interpretation. In contrast, the mixed-effects model incorporates random effects for tweet-level and language-level variability, showing that gender does not significantly influence labeling. This aligns with \citet{biester2022analyzing}, who found no significant gender-based differences in annotation behavior across various NLP tasks. Based on these findings, we use the mixed-effects model for further analysis, as it offers a more robust and accurate framework for interpreting demographic impacts.

\subsection{Random Effects Interpretation}
\label{subsec: Random_Effects}
The odds ratio (OR)\footnote{The odds ratio ($\text{OR} = e^{\beta}$) refers to how the odds of the outcome (\textit{label = yes}) change when a predictor variable changes, while all other variables are held constant.} for English tweets (OR = 0.84) indicates they are less likely to be labeled as sexist compared to Spanish tweets (OR = 1.95). Among the 347 annotators labeling Spanish tweets, 223 (64.27\%) are from Spanish-speaking countries, while only 73 out of 302 (24.17\%) annotators labeling English tweets are from English-speaking countries. Although we assume annotators are fluent in the language they label, regional residency may influence familiarity with cultural nuances and idiomatic expressions, affecting labeling decisions. Additionally, the grammatical structure of Spanish—being a gendered language—may make gender biases more explicit than in English. This aligns with \citet{lomotey2015sexism}, who emphasize the impact of grammatical gendering in Spanish. Thus, the observed differences in labeling may reflect both linguistic and cultural factors. Also, prior studies have found that classifiers achieve higher sexism-detection performance in English than in Spanish, likely due to the greater abundance of English-language training resources~\citep{fivez2024clin33}.

\subsection{Fixed Effects Interpretation}
\label{subsec: Fixed_Effects}
While the OR for females is slightly above 1, suggesting women may be more attuned to gender bias, gender does not significantly influence labeling decisions. Male and female annotators exhibit similar behavior, supported by a 74\% agreement in majority labeling, indicating consistency across genders. 
Similarly, although older annotators show slightly higher ORs, suggesting greater sensitivity to sexist content, no significant differences are observed across age groups, indicating age is not a decisive factor in labeling behavior.
In contrast, ethnicity significantly affects labeling. Black annotators are more likely to label tweets as sexist (OR = 5.50), while Latino annotators are less likely compared to White annotators (OR = 0.46). These findings align with \citet{tahaei2024analysis} and \citet{kwarteng2023understanding}, which highlight the heightened sensitivity of Black annotators, particularly Black women, due to lived experiences with intersectional discrimination. The lower likelihood among Latino annotators may reflect cultural norms.
Regarding education, no significant differences are found between annotators with bachelor’s and master’s degrees. However, those with only a high school degree are significantly less likely to label tweets as sexist (OR = 0.63).
Geographical location also plays a key role. Annotators from Africa are much less likely to label tweets as sexist (OR = 0.06), supporting findings from \citet{tahaei2024analysis} that emphasize the influence of country of origin and linguistic background on annotation behavior.

Our analysis shows that tweet-specific characteristics have a substantial impact on annotation outcomes, outweighing the influence of annotator demographics. While demographic features such as ethnicity, region, and education exhibit some significant associations with labeling tendencies, our mixed-effects model indicates that these effects are secondary to the inherent properties of the tweets. With an intraclass correlation coefficient (ICC) of 92\%, the majority of the variance in labeling outcomes is attributed to tweet-level variability, with language contributing only a minor additional source of variation. The remaining 8\% of the variance is explained by demographic variables and residual error. These findings suggest that, although demographic biases are not the dominant source of variability, they still play a meaningful role and should not be overlooked.

\vspace{-3pt}
\subsection{BERT Model Interpretation}
\label{subsec: BERT_Model_Interpretation}
We employed a multilingual BERT model for binary sexism classification, fine-tuning it on 90\% of the dataset using class weights and early stopping. Evaluated on the remaining 10\% (ensuring representation of all demographic combinations), the model achieved test accuracies of 77\% in English and 79\% in Spanish, demonstrating consistent cross-lingual performance.
To interpret the model's decisions, particularly for classifying tweets as sexist (YES), we utilized SHAP values. Calculating normalized mean SHAP importance for tokens in correctly classified YES instances revealed insights into feature attribution. 

\begin{figure}[H]
\centering
\includegraphics[width=0.4\textwidth]{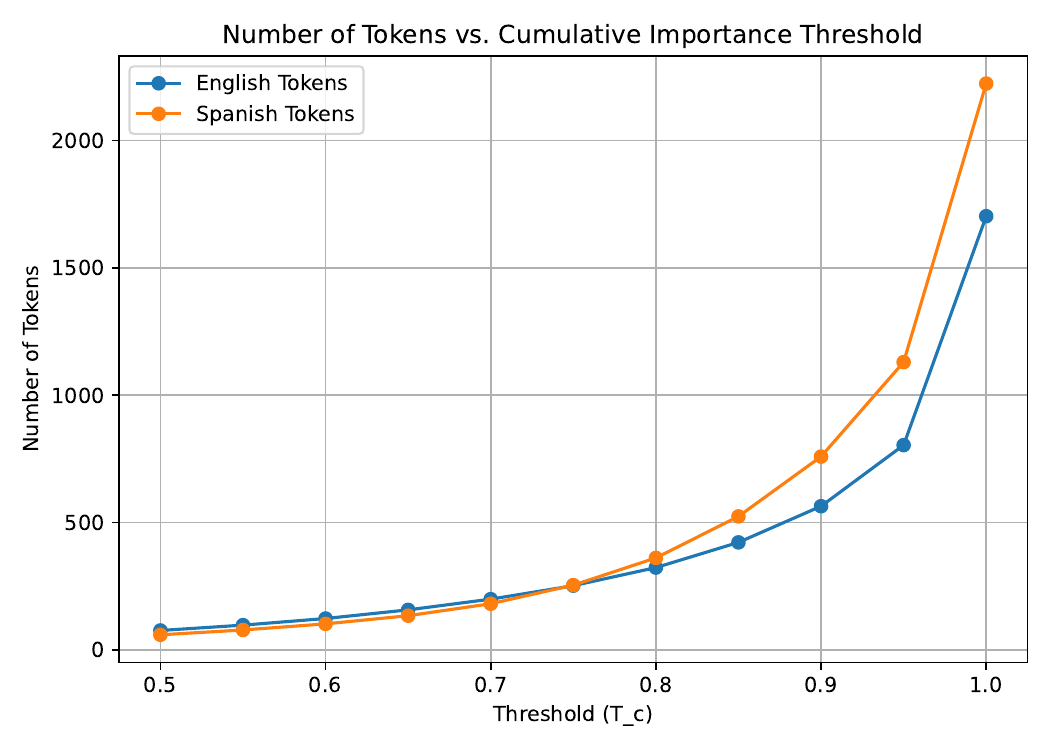}
\caption{\small Threshold vs. Number of Selected Tokens in both English and Spanish.}
\label{fig:threshold_vs_tokens}
\end{figure}

As shown in Figure \ref{fig:threshold_vs_tokens}, while a relatively small number of tokens capture roughly 50\% of the cumulative importance, explaining near-total importance (e.g., 95\%) necessitates considering a significantly larger lexicon, a trend particularly pronounced in Spanish. This suggests reliance on both core indicators and a broader range of terms for comprehensive detection. Examining the most influential tokens provides further clarity.
\vspace{-3pt}
\vspace{-12pt}

\begin{figure}[H]
\centering
% First plot
\begin{subfigure}[b]{0.4\textwidth}
\centering
\includegraphics[width=\textwidth]{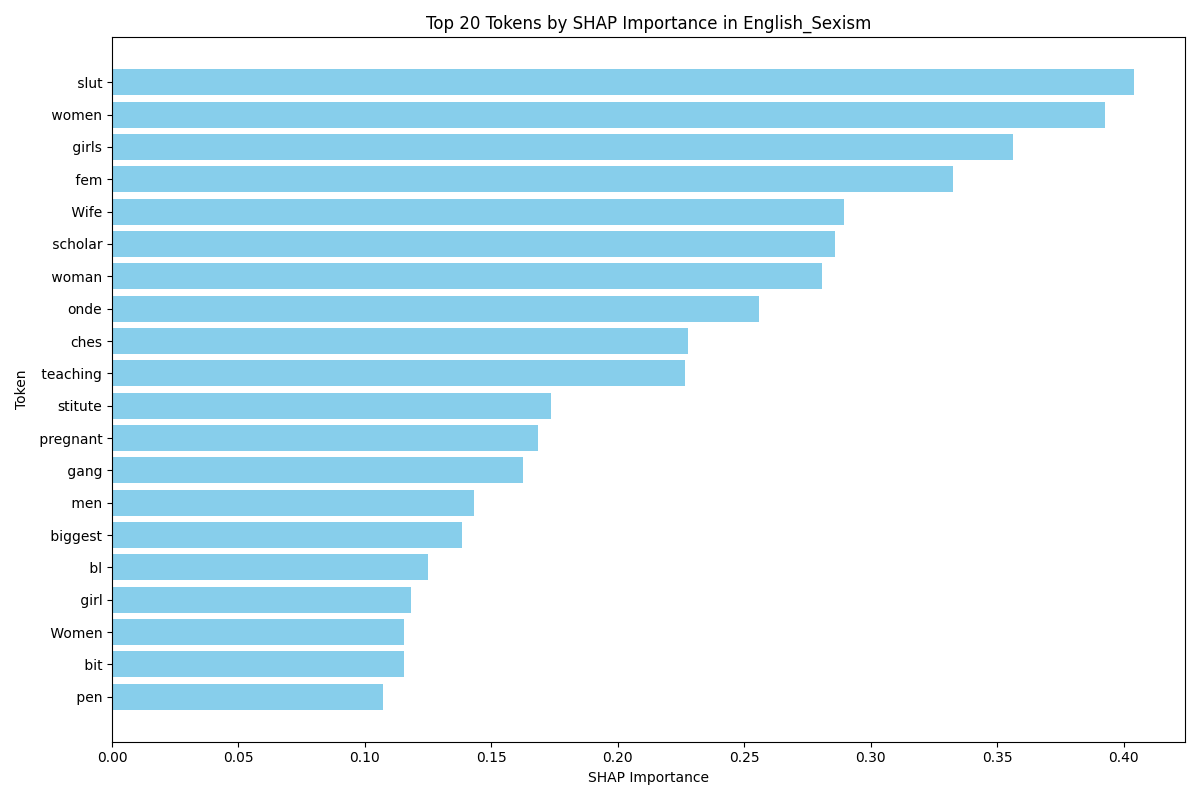}
\caption{\small English Tokens}
\label{fig:top_tokens_en}
\end{subfigure}%
\hspace{0.03\textwidth} % Adds small horizontal space between the two subfigures
% Second plot
\begin{subfigure}[b]{0.4\textwidth}
\centering
\includegraphics[width=\textwidth]{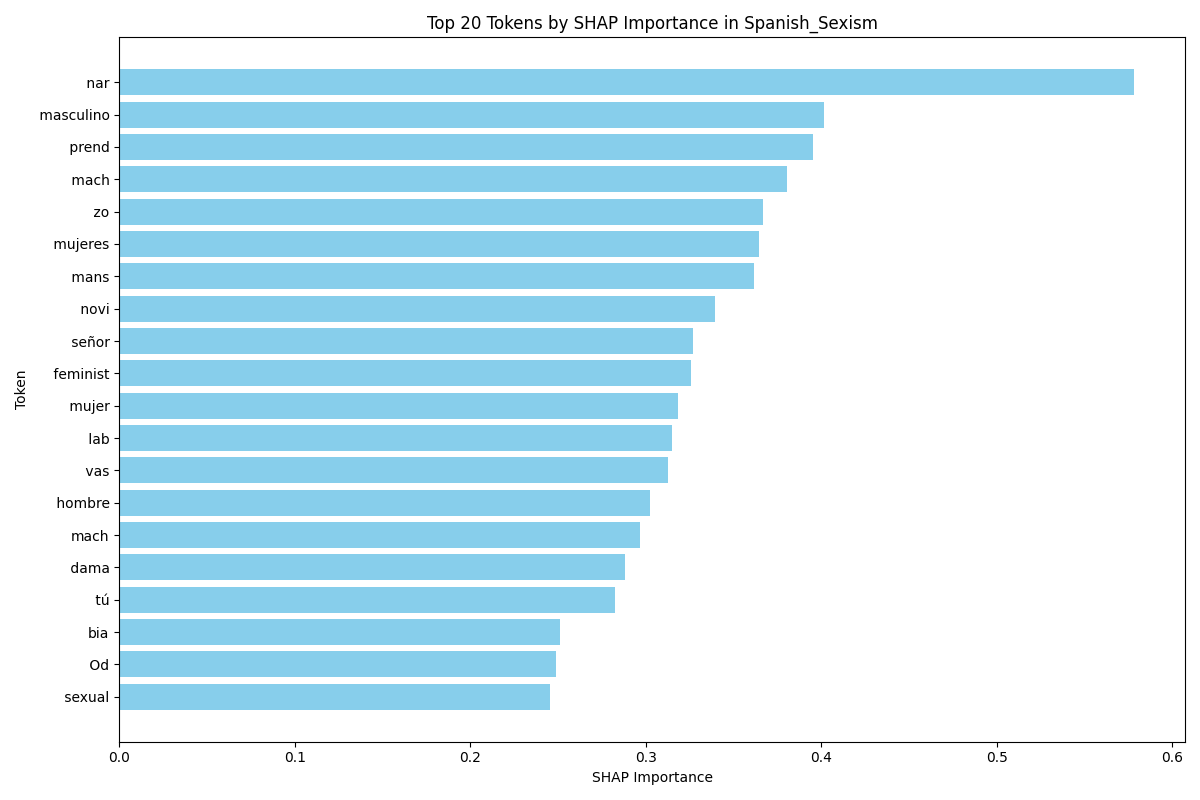}
\caption{\small Spanish Tokens}
\label{fig:top_tokens_es}
\end{subfigure}
\caption{\small Top 20 tokens by SHAP importance in (a) English and (b) Spanish.}
\label{fig:top_tokens}
\end{figure}

 Figure \ref{fig:top_tokens} displays the top 20 tokens by SHAP importance. In English, terms like \texttt{slut}, \texttt{women}, \texttt{girls}, and \texttt{wife} dominate, highlighting the model's focus on overtly gendered and potentially insulting language. Similarly, in Spanish, tokens such as \texttt{masculino}, \texttt{mujeres}, \texttt{feminist}, \texttt{mujer}, \texttt{mach}, and \texttt{sexual} are highly ranked, indicating a strong reliance on explicit gendered terms and references to sexual characteristics or ideologies.

\vspace{-3pt}
\vspace{-7pt}

 This analysis confirms that both language models heavily weigh content features directly related to sexism. While the specific influential tokens differ due to linguistic variations, the underlying mechanism points towards content-based classification. The distribution of influence also varies slightly, with the top 50 tokens accounting for ~40\% of importance in English versus ~45\% in Spanish (Appendix \ref{appendix:tokens}, \ref{appendix:Cumulative_importance}), suggesting a slightly more concentrated reliance on key terms in Spanish.\footnote{An exploratory analysis of unique token diversity across annotator demographic groups, detailed in Appendix \ref{appendix:unique_tokens_analysis}.} 
 
\subsection{GenAI Scenarios Results}
\label{subsec: GenAI_Scenarios_Results}
We evaluate our approach on a 10\% random sample of the dataset, comprising 326 English texts and 366 Spanish texts, covering all demographic groups. We measure performance using accuracy and F1-score. Table~\ref{tab:scenarios_metrics} presents the performance metrics for all GenAI scenarios across four models, LLaMA 3.2 3B, LLaMA 3.3 70B, OpenAI GPT-4o, and OpenAI GPT 4o-mini, in both English and Spanish. 

\vspace{-5pt}
\begin{table}[H]
\begin{scriptsize}
\caption{\small Performance metrics for all scenarios (see section \ref{subsec:GenAI Scenarios}). Numbers represent the scenarios: 1.GenAI, 2.GenP, 3.GenXAI, and 4.GenPXAI.}
\vspace{-5pt}
\label{tab:scenarios_metrics}
\centering
\renewcommand{\arraystretch}{1}
\begin{tabular}{l|p{0.35cm}p{0.35cm}p{0.35cm}p{0.35cm}|p{0.35cm}p{0.35cm}p{0.35cm}p{0.35cm}}
\toprule
\multirow{2}{*}{\textbf{Accuracy}} & \multicolumn{4}{c|}{\textbf{English}} & \multicolumn{4}{c}{\textbf{Spanish}} \\
\cline{2-9}
 & \textbf{1} & \textbf{2} & \textbf{3} & \textbf{4} & \textbf{1} & \textbf{2} & \textbf{3} & \textbf{4} \\
\midrule
% \multicolumn{9}{l}{\textbf{Accuracy}} \\

LM 3B     & 0.50 & 0.47 & \textbf{0.59} & 0.53 & 0.43 & 0.43 & 0.48 & \textbf{0.50} \\
LM 70B    & \textbf{0.66} & 0.64 & 0.65 & 0.64 & \textbf{0.64} & 0.58 & 0.57 & 0.58 \\
GPT-4o    & 0.76 & 0.75 & 0.73 & \textbf{0.78} & 0.75 & 0.77 & 0.72 & \textbf{0.77} \\
4o-mini   & \textbf{0.79} & 0.78 & 0.77 & \textbf{0.79} & 0.81 & 0.80 & \textbf{0.82} & 0.79 \\
\midrule
\multicolumn{9}{l}{\textbf{F1-score}} \\
\midrule
LM 3B     & 0.51 & 0.47 & 0.53 & \textbf{0.53} & 0.43 & 0.43 & 0.45 & \textbf{0.47} \\
LM 70B    & \textbf{0.66} & 0.60 & 0.62 & 0.58 & \textbf{0.62} & 0.51 & 0.49 & 0.47 \\
GPT-4o    & 0.74 & 0.74 & 0.71 & \textbf{0.77} & 0.74 & 0.76 & 0.70 & \textbf{0.76} \\
4o-mini   & 0.78 & 0.78 & 0.77 & \textbf{0.79} & 0.81 & 0.80 & \textbf{0.82} & 0.79 \\
\bottomrule
\end{tabular}
\end{scriptsize}
\end{table}
\vspace{-7pt}

Overall, OpenAI GPT 4o-mini and GPT-4o perform better, while LLaMA 3.2 3B tends to perform worse, and LLaMA 3.3 70B is in between. The English subset often shows a baseline advantage for the more capable models, while the Spanish subset sometimes benefits more from certain prompting strategies.
Differences across scenarios help reveal the impact of introducing personas and focusing attention on important tokens (XAI). Critically, assessing the utility of demographic personas (Scenario 2, GenP), we observe that it often provides no significant improvement over the baseline GenAI (Scenario 1) and occasionally leads to worse performance (e.g., LLaMA 3B and 70B models show decreased accuracy or F1-score in English, and LLaMA 70B sees a notable drop in F1-score in Spanish when personas are added). Even for the higher-performing GPT models, the gains from persona prompting alone are minimal or absent (e.g., GPT-4o mini accuracy slightly decreases in both languages). This suggests that simply layering demographic characteristics onto the prompt does not reliably enhance the LLM's ability to replicate nuanced human judgments for this task, questioning the value of such personas for improving annotation reliability.

Focusing on XAI (Scenario 3, GenXAI), highlighting important tokens identified by SHAP often helps smaller models (e.g., LLaMA 3.2 3B shows a marked improvement in accuracy in English going from 0.50 to 0.59, and in Spanish from 0.43 to 0.48) and provides a solid baseline, suggesting benefit from focusing the model on content features deemed important by an explainability analysis. For larger models, the effect of XAI alone is mixed, sometimes resulting in slight performance dips compared to the baseline (e.g., GPT-4o). For larger models, the combined approach (Scenario 4, GenPXAI) sometimes yields the highest scores (e.g., GPT-4o achieves its peak accuracy and F1 in both languages, and 4o-mini peaks in English). However, the improvement of GenPXAI over GenXAI is often marginal or inconsistent. For instance, with GPT-4o mini in Spanish, the GenXAI scenario (0.82 Acc, 0.82 F1) actually slightly outperforms the combined GenPXAI scenario (0.79 Acc, 0.79 F1). This pattern raises questions about whether the persona component in GenPXAI adds substantial value beyond the guidance provided by the content-focused XAI highlighting. The data suggests that directing the model's attention to relevant textual features (XAI) might be the more robust and impactful strategy, rather than attempting to simulate demographic perspectives through personas, whose contribution appears less certain. In summary, these results indicate that while baseline GenAI models already achieve strong performance on this task, the addition of demographic persona information offers questionable and inconsistent benefits for improving annotation reliability in this context. Guiding the model's attention using XAI based on content features appears more consistently helpful, particularly when paired with capable models, suggesting that focusing on the text itself through explainability methods is a more promising path forward than relying on potentially superficial persona simulation.

\section{Conclusion}
\label{sec:Conclusion}
This study evaluated the reliability of LLM annotations for sexism detection, focusing on the roles of annotator demographics and model explainability. Mixed-effects modeling showed that demographic factors, while sometimes statistically significant, accounted for only ~8\% of the variance in human labels, tweet content and individual differences were the main drivers. We tested the use of demographic personas to guide LLMs but found this strategy had limited, inconsistent, and sometimes negative effects on performance. SHAP analysis confirmed that content drove model decisions. These findings suggest that bias mitigation should focus less on broad demographic corrections and more on content and individual-level understanding. Simulated personas may oversimplify complexity and risk reinforcing stereotypes. This limitation is underscored by evidence that LLM often exhibits uniform stylistic patterns~\citep{mohammadi2025explainability}, showing that current models cannot fully emulate the diverse differences of human annotators. Instead, explainability tools that highlight 
content-relevant features offer a more promising path toward fairness and reliability in NLP.Future research should explore richer ways to capture diverse perspectives and improve content-based guidance in LLM annotations.

\section{Limitations and Future Work}

% Although our analysis suggests that demographic factors account for only a fraction of the variability in the labelling, our findings may not generalise to other languages or cultural contexts. Future work should examine a wider range of datasets and linguistic settings to better assess the robustness and cross-cultural applicability of our approach.

Although our analysis suggests that demographic factors account for only a fraction of the variability in the labeling, our findings may not generalise to other languages or cultural contexts. Future work should examine a wider range of datasets and linguistic settings to better assess the robustness and cross-cultural applicability of our approach.Our persona-driven prompts and explainability techniques rely on relatively broad demographic categories, which cannot capture the full richness of individual identities or personal experiences. Additionally, LLMs can exhibit hidden biases derivedfrom their training data, and our prompts may notalways surface or mitigate these biases.

%Although our analysis indicates that demographic factors account for only a fraction of labeling variability, this study concentrates on a single dataset (EXIST 2024) consisting of English and Spanish tweets. Consequently, the methods and findings may not generalize to other languages or cultural contexts. 

\section*{Reproducibility}
All codes and experiment notebooks are available on GitHub.\footnote{\url{https://github.com/mohammadi-hadi/Explainable_Annotations_Reliability}}

\section*{Acknowledgments}

We thank the organizers and annotators for their contributions in EXIST 2024 and acknowledge support from the focus area Applied Data Science (ADS) funding from Utrecht University.

\bibliography{custom}

\newpage
\appendix
\section{Annotators' demographic combination}
\label{appendix:demographiccombinations}

The total number of unique demographic combinations after removing those with rare representations.

\begin{table}[H]
\centering
\begin{scriptsize}
\caption{\small Unique Demographic Combinations}
\label{tab:unique_combinations}

\begin{tabular}{p{0.2cm}|p{4cm}p{0.7cm}|p{0.7cm}}
\toprule
\# & \textbf{Possible Combination} & \textbf{\# of es Ann} & \textbf{\# of en Ann} \\
\midrule
1 & F, 18-22, Black, Bachelor, Africa & 0 & 4 \\
2 & F, 18-22, Black, High school, Africa & 0 & 3 \\
3 & F, 18-22, Latino, Bachelor, America & 19 & 1 \\
4 & F, 18-22, Latino, High school, America & 15 & 4 \\
5 & F, 18-22, Latino, High school, Europe & 1 & 1 \\
6 & F, 18-22, White, Bachelor, America & 2 & 0 \\
7 & F, 18-22, White, Bachelor, Europe & 15 & 18 \\
8 & F, 18-22, White, High school, Europe & 7 & 25 \\
9 & F, 23-45, Black, Bachelor, Africa & 0 & 9 \\
10 & F, 23-45, Black, High school, Africa & 0 & 2 \\
11 & F, 23-45, Latino, Bachelor, America & 34 & 0 \\
12 & F, 23-45, Latino, High school, America & 6 & 0 \\
13 & F, 23-45, Latino, Master, America & 2 & 0 \\
14 & F, 23-45, White, Bachelor, America & 1 & 1 \\
15 & F, 23-45, White, Bachelor, Europe & 7 & 20 \\
16 & F, 23-45, White, High school, Europe & 1 & 3 \\
17 & F, 23-45, White, Master, Europe & 9 & 14 \\
18 & F, 46+, Black, Bachelor, Africa & 0 & 4 \\
19 & F, 46+, Latino, Bachelor, America & 12 & 0 \\
20 & F, 46+, Latino, Bachelor, Europe & 3 & 0 \\
21 & F, 46+, Latino, High school, America & 2 & 1 \\
22 & F, 46+, Latino, Master, America & 6 & 1 \\
23 & F, 46+, White, Bachelor, America & 3 & 2 \\
24 & F, 46+, White, Bachelor, Europe & 11 & 9 \\
25 & F, 46+, White, High school, Africa & 0 & 3 \\
26 & F, 46+, White, High school, America & 2 & 2 \\
27 & F, 46+, White, High school, Europe & 4 & 16 \\
28 & F, 46+, White, Master, America & 2 & 0 \\
29 & F, 46+, White, Master, Europe & 7 & 6 \\
30 & M, 18-22, Black, Bachelor, Africa & 0 & 2 \\
31 & M, 18-22, Black, High school, Africa & 0 & 2 \\
32 & M, 18-22, Latino, Bachelor, America & 10 & 2 \\
33 & M, 18-22, Latino, Bachelor, Europe & 1 & 2 \\
34 & M, 18-22, Latino, High school, America & 17 & 7 \\
35 & M, 18-22, Latino, High school, Europe & 3 & 2 \\
36 & M, 18-22, Latino, Master, Europe & 2 & 0 \\
37 & M, 18-22, White, Bachelor, Europe & 17 & 11 \\
38 & M, 18-22, White, High school, Europe & 11 & 25 \\
39 & M, 18-22, White, Master, Europe & 0 & 3 \\
40 & M, 23-45, Black, Bachelor, Africa & 0 & 7 \\
41 & M, 23-45, Black, Master, Africa & 0 & 2 \\
42 & M, 23-45, Latino, Bachelor, America & 8 & 5 \\
43 & M, 23-45, Latino, Bachelor, Europe & 1 & 2 \\
44 & M, 23-45, Latino, Master, America & 2 & 0 \\
45 & M, 23-45, Latino, Master, Europe & 2 & 0 \\
46 & M, 23-45, White, Bachelor, Europe & 24 & 10 \\
47 & M, 23-45, White, High school, Europe & 4 & 10 \\
48 & M, 23-45, White, Master, Europe & 18 & 15 \\
49 & M, 46+, Latino, Bachelor, America & 8 & 3 \\
50 & M, 46+, Latino, Master, America & 2 & 0 \\
51 & M, 46+, White, Bachelor, Africa & 0 & 2 \\
52 & M, 46+, White, Bachelor, America & 5 & 5 \\
53 & M, 46+, White, Bachelor, Europe & 21 & 14 \\
54 & M, 46+, White, High school, America & 0 & 2 \\
55 & M, 46+, White, High school, Europe & 12 & 15 \\
56 & M, 46+, White, Master, Europe & 8 & 5 \\
\bottomrule
\end{tabular}
\end{scriptsize}

\end{table}

\section{Unique Token Analysis by Demographic Group}
\label{appendix:unique_tokens_analysis}

\begin{figure}[ht]
    \centering
    % First plot
    \begin{subfigure}[b]{0.4\textwidth}
        \centering
        \includegraphics[width=\textwidth]{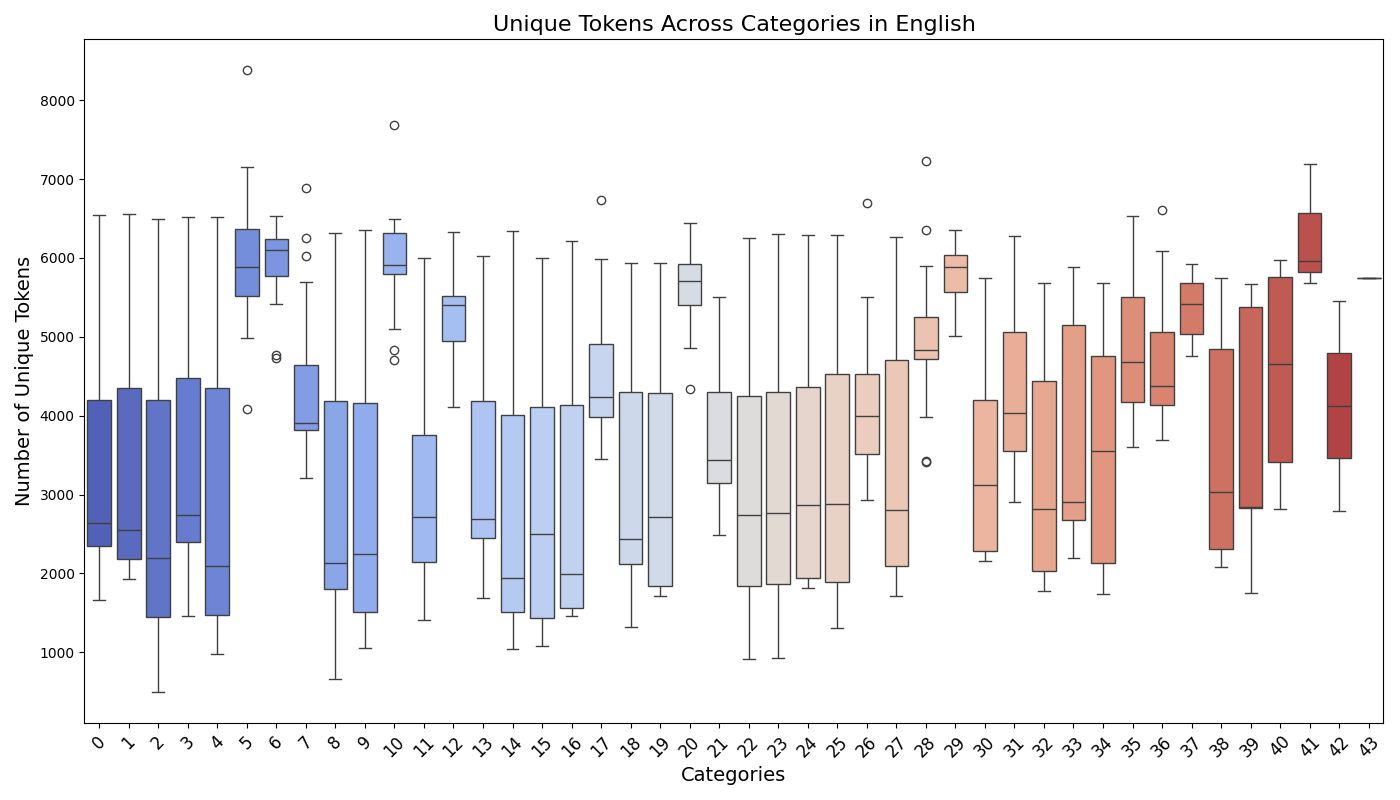}
        \caption{\small English Tokens}
        \label{fig:appendix_unique_tokens_en}
    \end{subfigure}%
    \hspace{0.05\textwidth} % Adds small horizontal space between the two subfigures
    % Second plot
    \begin{subfigure}[b]{0.4\textwidth}
        \centering
        \includegraphics[width=\textwidth]{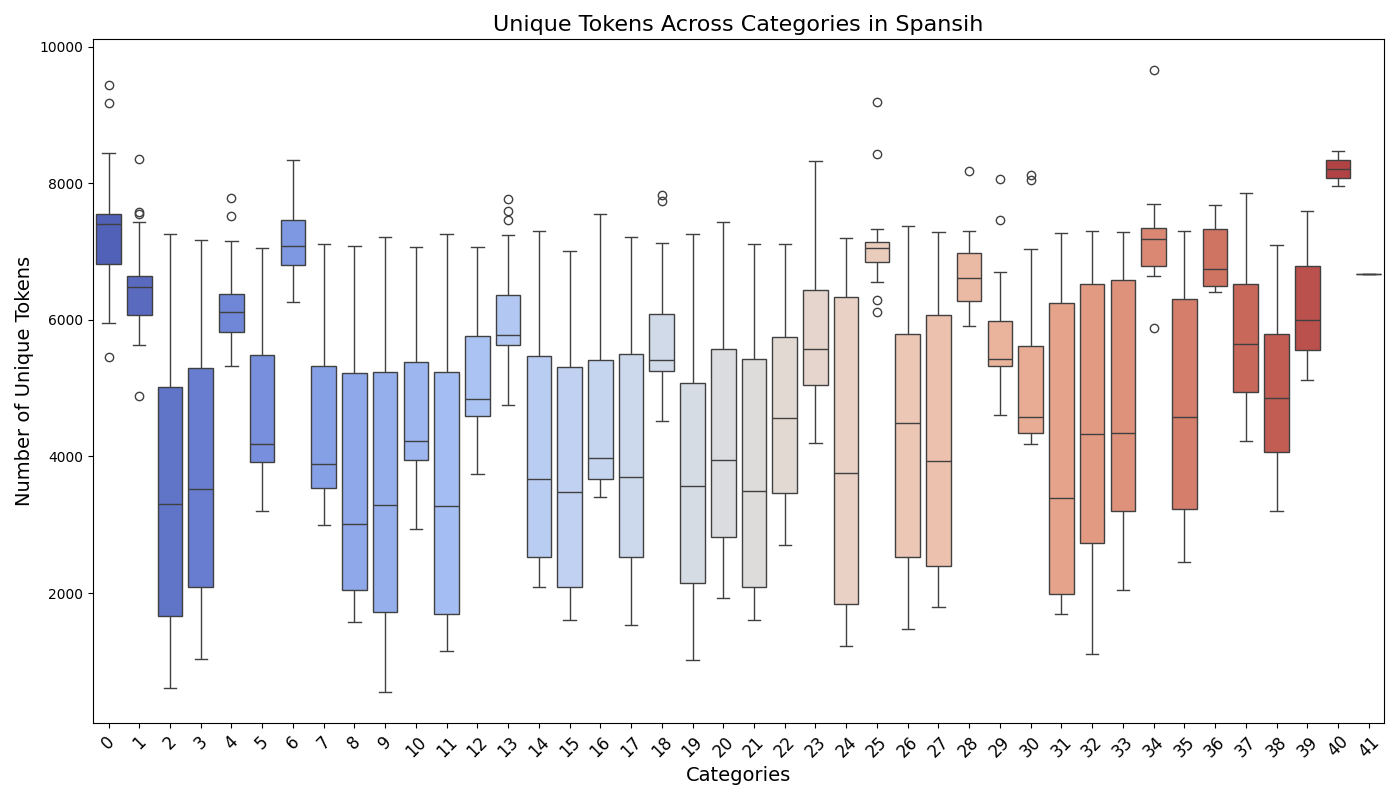}
        \caption{\small Spanish Tokens}
        \label{fig:appendix_unique_tokens_es}
    \end{subfigure}
    \caption{\small Distribution of unique tokens per tweet across various annotator demographic categories in (a) English and (b) Spanish. This exploratory analysis hints at subtle variations but confirms the secondary nature of these effects compared to content.}
    \label{fig:appendix_unique_tokens} % Note: New label for the appendix figure
\end{figure}

To further explore potential secondary demographic influences, we analyzed the distribution of unique token counts within tweets annotated by different demographic groups. This exploratory analysis aimed to identify potential variations in linguistic engagement or lexical diversity associated with annotator characteristics. As shown in Figure \ref{fig:appendix_unique_tokens}, we observed some variation across categories in both English and Spanish. For instance, certain groups exhibited broader ranges of unique tokens, potentially hinting at subtle cultural or linguistic factors influencing how they engage with the text. However, consistent with our primary findings, these observed differences appear secondary to the overwhelming influence of the tweet content itself on the annotation process and model interpretation.

\section{Top ten demographic combination}
\label{appendix:demographic_combination}

\begin{figure}[H]
    \centering
        \centering
        \includegraphics[width=0.4\textwidth]{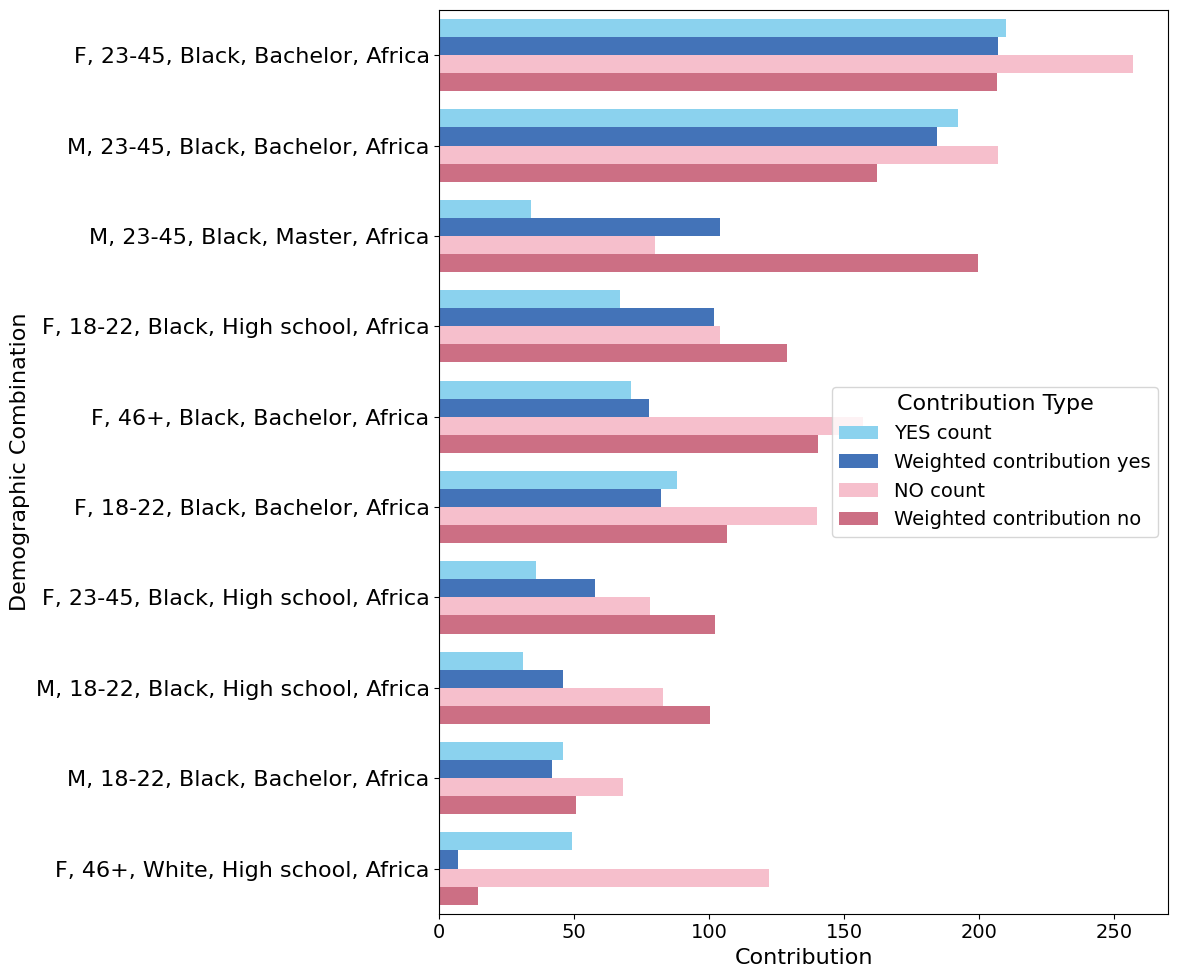}
        \caption{\small Different demographic combinations that have the highest weight contributions across both label classes}
        
        \label{fig:weight_contribution}
\end{figure}%

\section{Complete Lists of Important Tokens}
\label{appendix:tokens}

Here are the tokens identified by the model that contribute to classifying tweets as sexist, along with their importance scores. 

\renewcommand{\arraystretch}{1.2}
\begin{table}[H]
\centering
\scriptsize
\caption{\small 50 Top important English Tokens}
\label{tab:en_tokens}
\begin{tabular}{p{0.6cm}p{0.5cm}p{0.5cm}p{0.7cm}|p{0.6cm}p{0.5cm}p{0.5cm}p{0.7cm}}
\toprule
\textbf{Token} & \textbf{SHAP} & \textbf{Ratio} & \textbf{Cum.} & \textbf{Token} & \textbf{SHAP} & \textbf{Ratio} & \textbf{Cum.} \\
\midrule
 slut & 0.4041 & 0.0246 & 0.0246 & feminist & 0.1017 & 0.0062 & 0.2818 \\
 women & 0.3928 & 0.0239 & 0.0485 & periods & 0.0991 & 0.0060 & 0.2878 \\
 girls & 0.3561 & 0.0217 & 0.0702 & pro & 0.0974 & 0.0059 & 0.2938 \\
 fem & 0.3324 & 0.0202 & 0.0905 & her & 0.0972 & 0.0059 & 0.2997 \\
 Wife & 0.2896 & 0.0176 & 0.1082 & ok & 0.0935 & 0.0057 & 0.3054 \\
 scholar & 0.2858 & 0.0174 & 0.1256 & She & 0.0924 & 0.0056 & 0.3110 \\
 woman & 0.2807 & 0.0171 & 0.1427 & boys & 0.0896 & 0.0054 & 0.3165 \\
 onde & 0.2559 & 0.0156 & 0.1583 & ti & 0.0871 & 0.0053 & 0.3218 \\
 ches & 0.2278 & 0.0138 & 0.1722 & Like & 0.0853 & 0.0052 & 0.3270 \\
 teaching & 0.2264 & 0.0138 & 0.1860 & mbo & 0.0837 & 0.0051 & 0.3321 \\
 stitute & 0.1735 & 0.0105 & 0.1966 & ips & 0.0836 & 0.0051 & 0.3372 \\
 pregnant & 0.1682 & 0.0102 & 0.2068 & ts & 0.0820 & 0.0050 & 0.3422 \\
 gang & 0.1624 & 0.0099 & 0.2167 & coverage & 0.0808 & 0.0049 & 0.3472 \\
 men & 0.1430 & 0.0087 & 0.2255 & really & 0.0806 & 0.0049 & 0.3521 \\
 biggest & 0.1382 & 0.0084 & 0.2339 & wife & 0.0776 & 0.0047 & 0.3568 \\
 bl & 0.1249 & 0.0076 & 0.2415 & dies & 0.0773 & 0.0047 & 0.3615 \\
 girl & 0.1182 & 0.0072 & 0.2487 & finger & 0.0768 & 0.0046 & 0.3662 \\
 Women & 0.1156 & 0.0070 & 0.2558 & trophy & 0.0759 & 0.0046 & 0.3708 \\
 bit & 0.1155 & 0.0070 & 0.2628 & dressed & 0.0747 & 0.0045 & 0.3754 \\
 pen & 0.1073 & 0.0065 & 0.2694 & ina & 0.0742 & 0.0045 & 0.3799 \\
 financial & 0.1021 & 0.0062 & 0.2756 & Why & 0.0739 & 0.0045 & 0.3844 \\
 female & 0.0734 & 0.0044 & 0.3889 & comment & 0.0733 & 0.0044 & 0.3934 \\
 dress & 0.0702 & 0.0042 & 0.3977 & sex & 0.0672 & 0.0041 & 0.4017 \\
 male & 0.0669 & 0.0040 & 0.4058 & husband & 0.0668 & 0.0040 & 0.4099\\
 ehan & 0.0654 & 0.0039 & 0.4139 & ouse & 0.0649 & 0.0039 & 0.4179 \\
\bottomrule
\end{tabular}
\end{table}

\renewcommand{\arraystretch}{1.2}
\begin{table}[H]
\centering
\scriptsize
\caption{\small 50 Top important Spanish Tokens}
\label{tab:es_tokens}
\begin{tabular}{p{0.7cm}p{0.5cm}p{0.5cm}p{0.7cm}|p{0.6cm}p{0.5cm}p{0.5cm}p{0.7cm}}
\toprule
\textbf{Token} & \textbf{SHAP} & \textbf{Ratio} & \textbf{Cum.} & \textbf{Token} & \textbf{SHAP} & \textbf{Ratio} & \textbf{Cum.} \\
\midrule
 apa & 0.1573 & 0.0063 & 0.3787 & feminist & 0.3258 & 0.0132 & 0.1557 \\
 ones & 0.1489 & 0.0060 & 0.3848 & mujer & 0.3184 & 0.0129 & 0.1686 \\
 ios & 0.1478 & 0.0059 & 0.3907 & lab & 0.3151 & 0.0127 & 0.1814 \\
 var & 0.1476 & 0.0059 & 0.3967 & vas & 0.3123 & 0.0126 & 0.1941 \\
 novia & 0.1416 & 0.0057 & 0.4025 & hombre & 0.3026 & 0.0122 & 0.2063 \\
 bian & 0.1415 & 0.0057 & 0.4082 & mach & 0.2965 & 0.0120 & 0.2184 \\
 golf & 0.1414 & 0.0057 & 0.4140 & dama & 0.2881 & 0.0116 & 0.2301 \\
 male & 0.1393 & 0.0056 & 0.4196 & t\'u & 0.2822 & 0.0114 & 0.2415 \\
 marido & 0.1384 & 0.0056 & 0.4252 & bia & 0.2508 & 0.0101 & 0.2517 \\
 tant & 0.1289 & 0.0052 & 0.4305 & Od & 0.2485 & 0.0100 & 0.2618 \\
 laga & 0.1269 & 0.0051 & 0.4356 & sexual & 0.2453 & 0.0099 & 0.2717 \\
 \~nas & 0.1242 & 0.0050 & 0.4406 & fem & 0.2309 & 0.0093 & 0.2811 \\
 ellas & 0.1235 & 0.0050 & 0.4457 & femenino & 0.2263 & 0.0091 & 0.2903 \\
 amo & 0.1227 & 0.0049 & 0.4506 & doctor & 0.2237 & 0.0090 & 0.2993 \\
 aca & 0.1179 & 0.0047 & 0.4554 & princesa & 0.2231 & 0.0090 & 0.3084 \\
 loc & 0.1080 & 0.0043 & 0.4598 & nen & 0.2200 & 0.0089 & 0.3173 \\
 ball & 0.1023 & 0.0041 & 0.4640 & masculin & 0.2189 & 0.0088 & 0.3262 \\
 nar & 0.5781 & 0.0234 & 0.0234 & Mujeres & 0.2137 & 0.0086 & 0.3349 \\
 masculino & 0.4012 & 0.0162 & 0.0397 & ni\~na & 0.2028 & 0.0082 & 0.3431 \\
 prend & 0.3953 & 0.0160 & 0.0557 & bella & 0.1890 & 0.0076 & 0.3508 \\
 mach & 0.3804 & 0.0154 & 0.0712 & ton & 0.1839 & 0.0074 & 0.3582 \\
 zo & 0.3665 & 0.0148 & 0.0860 & ni\~nos & 0.1807 & 0.0073 & 0.3656 \\
 mujeres & 0.3642 & 0.0147 & 0.1008 & ment & 0.1670 & 0.0067 & 0.3723 \\
 mans & 0.3615 & 0.0146 & 0.1155 & novi & 0.3394 & 0.0137 & 0.1292\\
 señor & 0.3266 & 0.0132 & 0.1425 &  sÃ & 0.1003 & 0.0040 & 0.4680 \\

\bottomrule
\end{tabular}
\end{table}

\section{Cumulative importance of the top 50 tokens}
\label{appendix:Cumulative_importance}

\begin{figure}[H]
        \centering
        \includegraphics[width=0.4\textwidth]{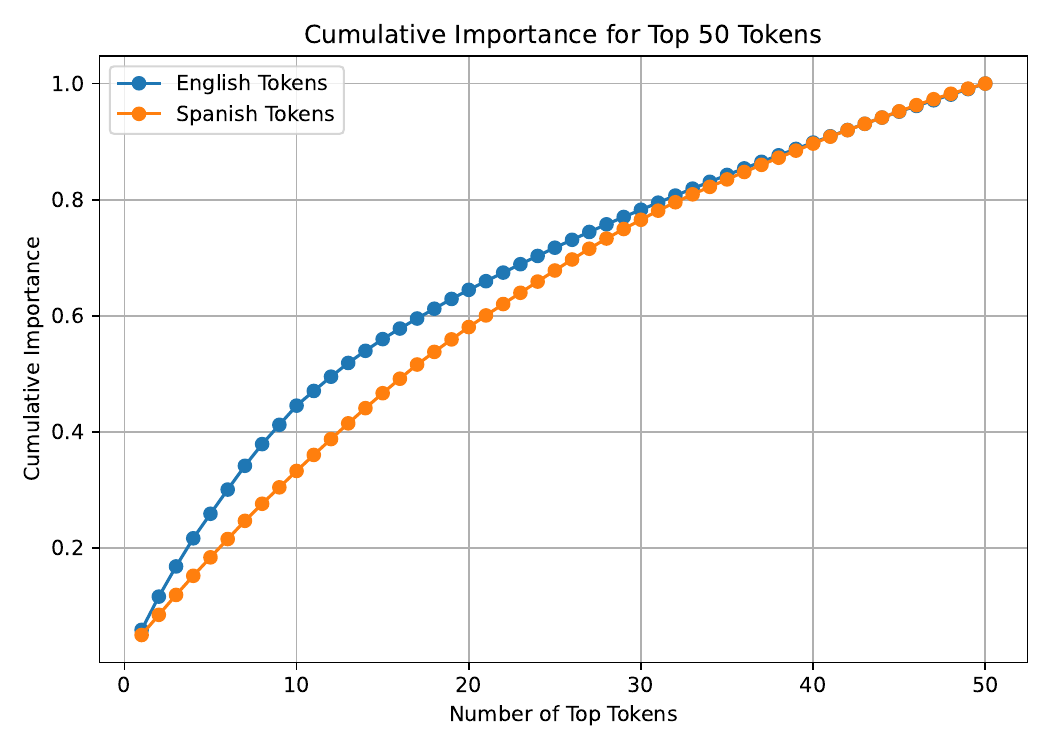}
        \caption{\small The cumulative importance of the top 50 tokens in both English and Spanish.}
        \label{fig:cumulative_importance}
\end{figure}

\section{LLMs Performance Comparison}
\label{appendix:performance_comparison}

\begin{figure}[H]
    \centering
    % First plot
    \begin{subfigure}[b]{0.4\textwidth}
        \centering
        \includegraphics[width=\textwidth]{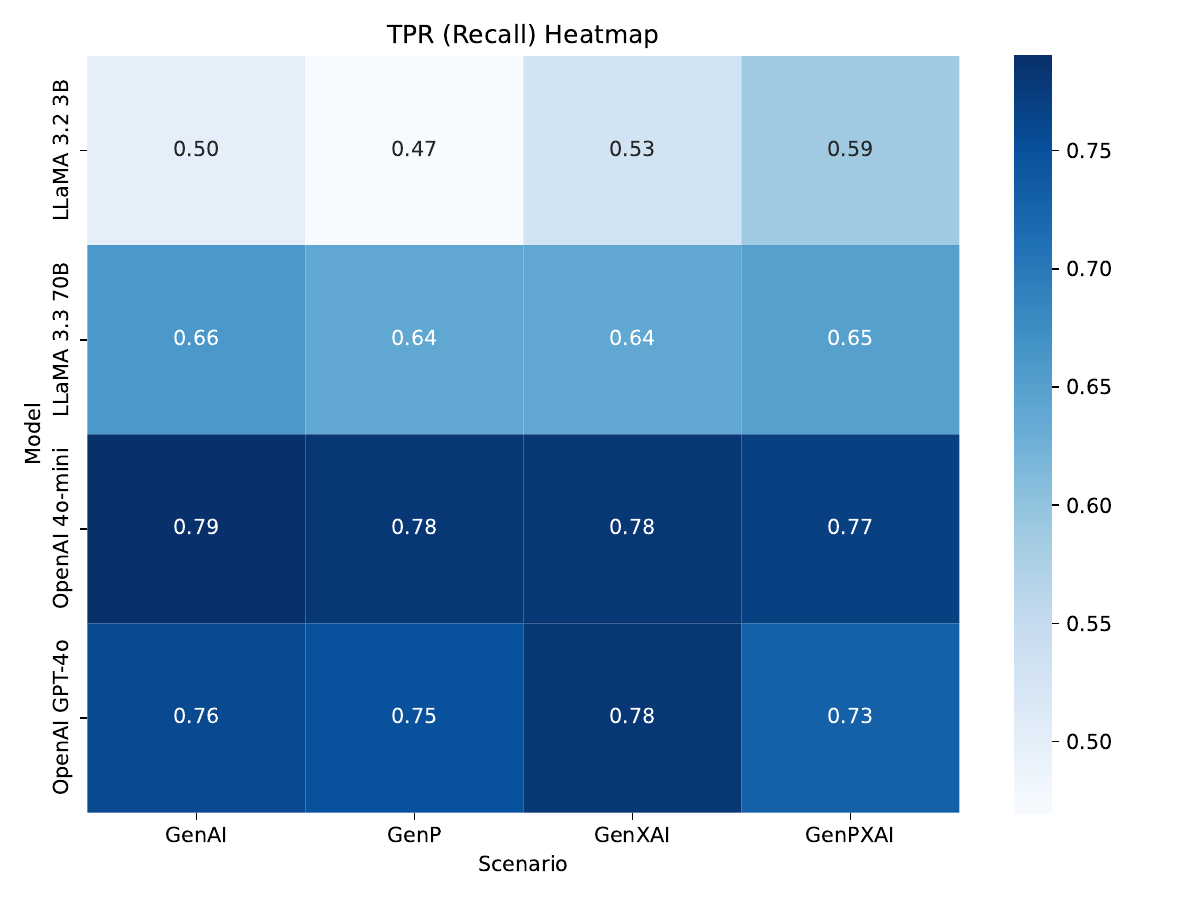}
        \caption{\small English}
        \label{fig:heatmap_en}
    \end{subfigure}%
    \hspace{0.05\textwidth} % Adds small horizontal space between the two subfigures
    % Second plot
    \begin{subfigure}[b]{0.4\textwidth}
        \centering
        \includegraphics[width=\textwidth]{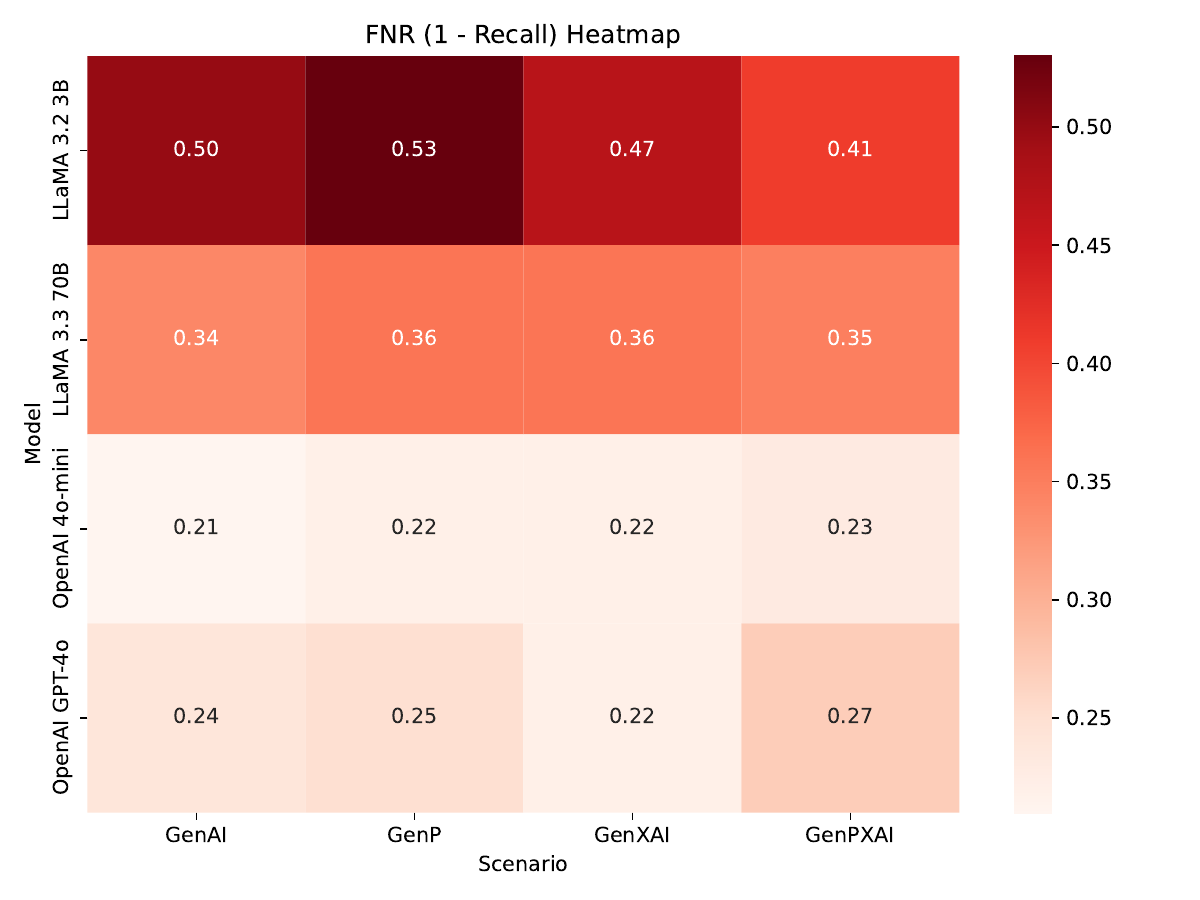}
        \caption{\small Spanish}
        \label{fig:heatmap_es}
    \end{subfigure}
    \caption{\small Comapring TPR and FNR across models, scenarios, and languages.}
    \label{fig:heatmap}
\end{figure}

\end{document}